\def\year{2018}\relax
\documentclass[letterpaper]{article}
\usepackage{aaai18}
\usepackage{times,helvet,courier}
\usepackage{graphicx}
\usepackage[tight,footnotesize]{subfigure}
\usepackage{amsmath,amssymb,amsthm,bm}
\usepackage{multirow,makecell,footmisc,url}
\usepackage[ruled,vlined]{algorithm2e}
\usepackage{color}

\begin{document}

\title{Transfer Adversarial Hashing for Hamming Space Retrieval}
\author{
   Zhangjie Cao, Mingsheng Long\thanks{Corresponding author: Mingsheng Long}, Chao Huang, Jianmin Wang\\
   KLiss, MOE; NEL-BDS; TNList; School of Software, Tsinghua University, China\\
   {\tt\small \{caozhangjie14,huangcthu\}@gmail.com \ \{mingsheng,jimwang\}@tsinghua.edu.cn}
}

\maketitle
\begin{abstract}
\begin{quote}
Hashing is widely applied to large-scale image retrieval due to the storage and retrieval efficiency. Existing work on deep hashing assumes that the database in the target domain is identically distributed with the training set in the source domain. This paper relaxes this assumption to a transfer retrieval setting, which allows the database and the training set to come from different but relevant domains. However, the transfer retrieval setting will introduce two technical difficulties: first, the hash model trained on the source domain cannot work well on the target domain due to the large distribution gap; second, the domain gap makes it difficult to concentrate the database points to be within a small Hamming ball. As a consequence, transfer retrieval performance within Hamming Radius 2 degrades significantly in existing hashing methods. This paper presents Transfer Adversarial Hashing (TAH), a new hybrid deep architecture that incorporates a pairwise $t$-distribution cross-entropy loss to learn concentrated hash codes and an adversarial network to align the data distributions between the source and target domains. TAH can generate compact transfer hash codes for efficient image retrieval on both source and target domains. Comprehensive experiments validate that TAH yields state of the art Hamming space retrieval performance on standard datasets.
\end{quote}
\end{abstract}

\section{Introduction}
With increasing large-scale and high-dimensional image data emerging in search engines and social networks, image retrieval has attracted increasing attention in computer vision community. Approximate nearest neighbors (ANN) search is an important method for image retrieval. Parallel to the traditional indexing methods \cite{cite:TOMM06CBIR}, another advantageous solution is hashing methods \cite{cite:Arxiv14HashSurvey}, which transform high-dimensional image data into compact binary codes and generate similar binary codes for similar data items. In this paper, we will focus on data-dependent hash encoding schemes for efficient image retrieval, which have shown better performance than data-independent hashing methods, e.g. Locality-Sensitive Hashing (LSH) \cite{cite:VLDB99LSH}.

There are two related search problems in hashing~\cite{cite:TPAMI14FES}, $K$-NN search and Point Location in Equal Balls (PLEB)~\cite{cite:STOC98ANN}. Given a database of hash codes, $K$-NN search aims to find $K$ codes in database that are closest in Hamming distance to a given query. With the Definition that a binary code is an $r$-$neighbor$ of a query code $q$ if it differs from $q$ in $r$ bits or less, PLEB for $r$ Equal Ball finds all $r$-$neighbors$ of a query in the database. This paper will focus on PLEB search which we call Hamming Space Retrieval. 

For binary codes of $b$ bits, the number of distinct hash buckets to examine is $N\left( {b,r} \right) = \sum\nolimits_{k = 0}^r {\left( {_k^b} \right)}$. $N\left( {b,r} \right)$ grows rapidly with $r$ and when $r\le2$, it only requires $O(1)$ time for each query to find all $r$-$neighbors$. Therefore, the search efficiency and quality within Hamming Radius 2 is an important technical backbone of hashing.

Previous image hashing methods~\cite{cite:NIPS09BRE,cite:CVPR11ITQ,cite:ICML11MLH,cite:CVPR12MIH,cite:CVPR12KSH,cite:TPAMI12SSH,cite:CVPR13HBS,cite:CVPR13BP,cite:ICML14CBE,cite:SIGIR14LFH,cite:CVPR14CH,cite:AAAI14CNNH,cite:CVPR15DNNH,cite:CVPR15SDH,cite:CVPR15DH,cite:AAAI16DHN,cite:AAAI16DQN,cite:IJCAI16DPSH,cite:CVPR2016DSH,cite:ICCV17HashNet} have achieved promising image retrieval performance. However, they all require that the source domain and the target domain are the same, under which they can directly apply the model trained on train images to database images. Many real-world applications actually violate this assumption where source and target domain are different. For example, one person want to build a search engine on real-world images, but unfortunately, he/she only has images rendered from 3D model with known similarity and real-world images without any supervised similarity. Thus, a method for the transfer setting is needed.

\begin{figure}[tbp]
  \centering
    \subfigure[DHN]{
        \includegraphics[width=0.13\textwidth]{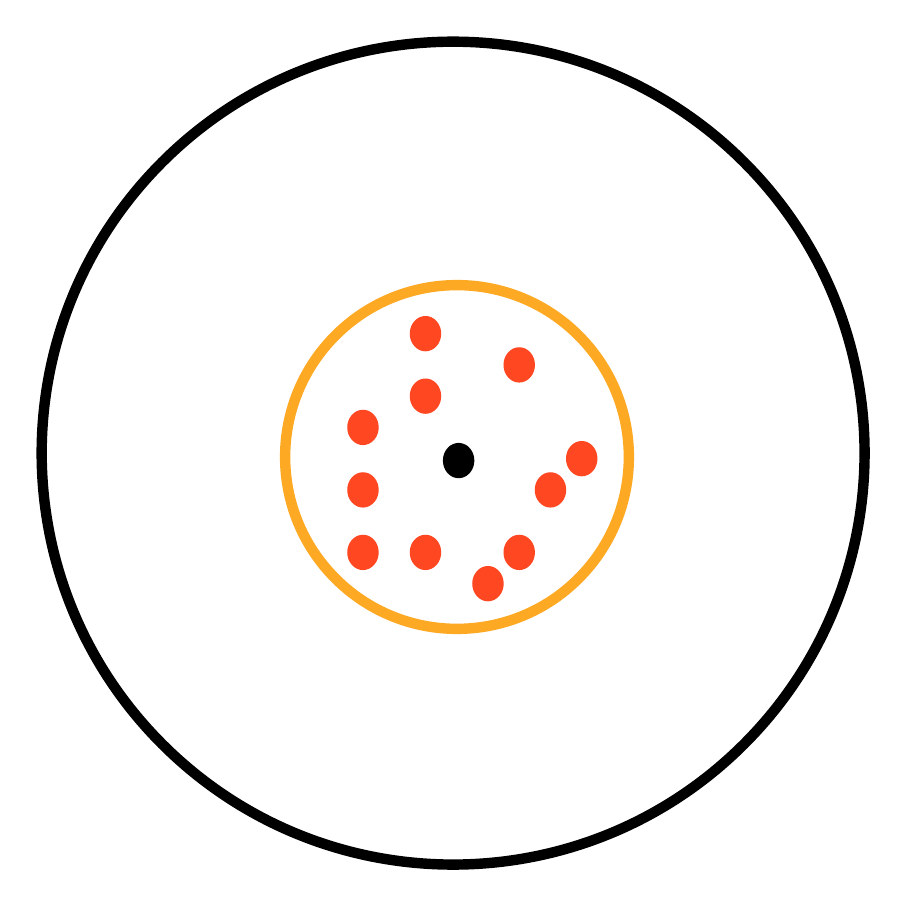}
        \label{fig:dhn_same}
    }
    \subfigure[DHN-Transfer]{
        \includegraphics[width=0.13\textwidth]{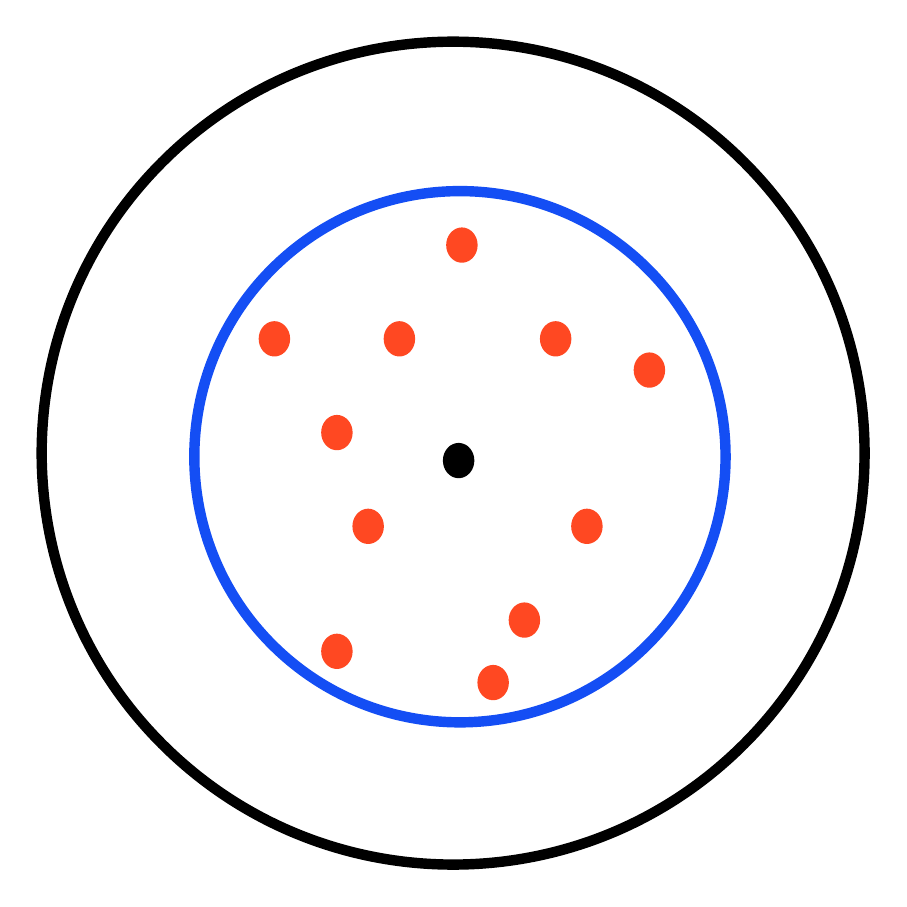}
        \label{fig:dhn_transfer}
    }
    \subfigure[$t$-Transfer]{
        \includegraphics[width=0.13\textwidth]{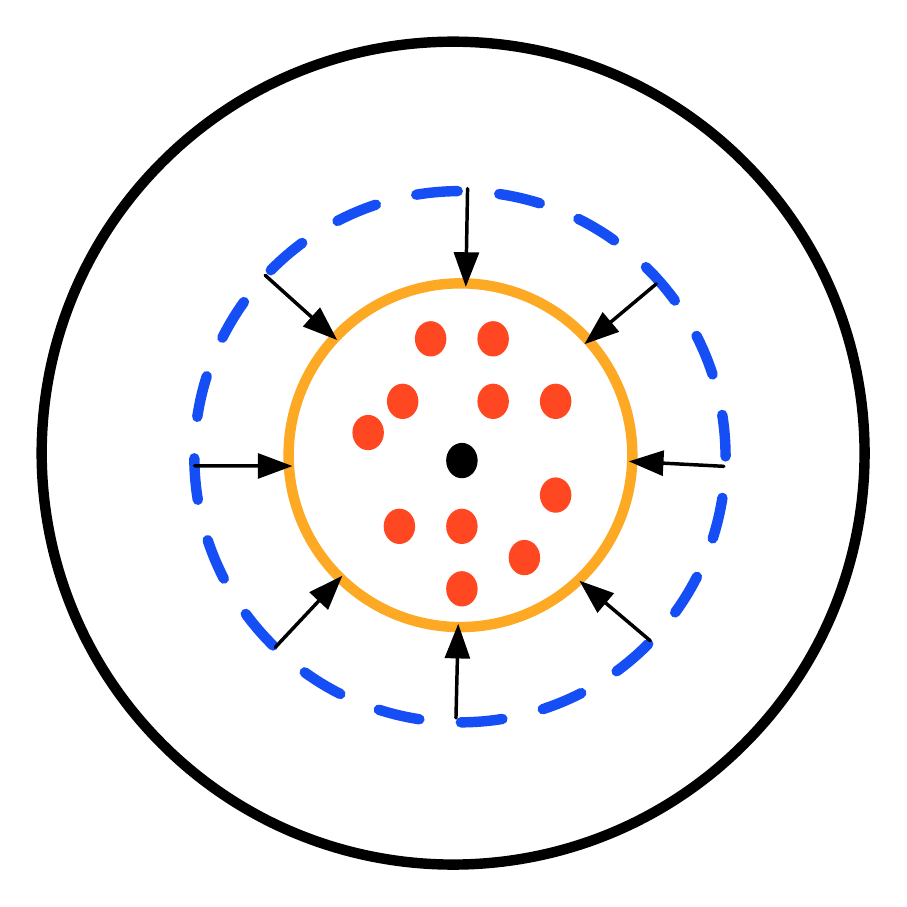}
        \label{fig:t_transfer}
    }
  \vspace{-10pt}
  \caption{Visualization of similar points within Hamming Radius 2 of a query.}
   \label{fig:problem}
   \vspace{-10pt}
\end{figure}

The transfer retrieval setting can raise two problems. The first is that the similar points of a query within its Hamming Radius 2 Ball will deviate more from the query. As shown in Figure~\ref{fig:dhn_same}, the red points similar to black query in the orange Hamming Ball (Hamming Radius 2 Ball) of the source domain scatter more sparsely in a blue larger Hamming Ball of the target domain in Figure~\ref{fig:dhn_transfer}, indicating that the number of similar points within Hamming Radius 2 decreases because of the domain gap. This can be validated in Table~\ref{table:problem_table} by the decreasing of average number of similar points of DHN from $1450$ on $synthetic \rightarrow synthetic$ task to $58$ on $real \rightarrow synthetic$ task. Thus, we propose a new similarity function based on $t$-distribution and Hamming distance, denoted as $t$-Transfer in Figure~\ref{fig:problem} and Table~\ref{table:problem_table}. From Figure~\ref{fig:dhn_transfer}-\ref{fig:t_transfer} and Table~\ref{table:problem_table}, we can observe that our proposed similarity function can draw similar points closer and let them locate in the Hamming Radius 2 Ball of the query.

\begin{table}[tbp]
    \centering 
    \addtolength{\tabcolsep}{0pt}
    \vspace{-10pt}
    \caption{Average Number of Similar Points within Hamming Radius 2 of each query on $synthetic \rightarrow synthetic$ and $real \rightarrow synthetic$ tasks in VisDA2017 dataset.}
    \label{table:problem_table}
    \begin{tabular}{c|c|c|c}
        \Xhline{1.0pt}
        Task & DHN  & DHN-Transfer & $t$-Transfer \\
        \hline
        \#Similar Points & 1450 & 58 & 620 \\
        \Xhline{1.0pt}
    \end{tabular}
\end{table}

The second problem is that substantial gap across Hamming spaces exists between source domain and target domain since they follow different distributions. We need to close this distribution gap. This paper exploits adversarial learning~\cite{cite:ICML15RevGrad} to align the distributions of source domain and target domain, to adapt the hashing model trained on source domain to target domain. With this domain distribution alignment, we can apply the hashing model trained on source domain to the target domain.

In all, this paper proposes a novel Transfer Adversarial Hashing (TAH) approach to the transfer setting for image retrieval. With similarity relationship learning and domain distribution alignment, we can align different domains in Hamming space and concentrate the hash codes to be within a small Hamming ball in an end-to-end deep architecture to enable efficient image retrieval within Hamming Radius 2. Extensive experiments show that TAH yields state of the art performance on public benchmarks NUS-WIDE and VisDA2017.

\section{Related Work}
Our work is related to learning to hash methods for image retrieval, which can be organized into two categories: unsupervised hashing and supervised hashing. We refer readers to \cite{cite:Arxiv14HashSurvey} for a comprehensive survey.

Unsupervised hashing methods learn hash functions that encode data points to binary codes by training from unlabeled data. Typical learning criteria include reconstruction error minimization \cite{cite:AI07SemanticHashing,cite:CVPR11ITQ,cite:TPAMI11PQ} and graph learning\cite{cite:NIPS09SH,cite:ICML11AGH}. While unsupervised methods are more general and can be trained without semantic labels or relevance information, they are subject to the semantic gap dilemma \cite{cite:TPAMI00SemanticGap} that high-level semantic description of an object differs from low-level feature descriptors. Supervised methods can incorporate semantic labels or relevance information to mitigate the semantic gap and improve the hashing quality significantly. 
Typical supervised methods include Binary Reconstruction Embedding (BRE) \cite{cite:NIPS09BRE}, Minimal Loss Hashing (MLH) \cite{cite:ICML11MLH} and Hamming Distance Metric Learning \cite{cite:NIPS12HDML}.
Supervised Hashing with Kernels (KSH) \cite{cite:CVPR12KSH} generates hash codes by minimizing the Hamming distances across similar pairs and maximizing the Hamming distances across dissimilar pairs.

As various deep convolutional neural networks (CNN) \cite{cite:NIPS12CNN,cite:CVPR16DRL} yield breakthrough performance on many computer vision tasks, deep learning to hash has attracted attention recently. CNNH \cite{cite:AAAI14CNNH} adopts a two-stage strategy in which the first stage learns hash codes and the second stage learns a deep network to map input images to the hash codes. DNNH \cite{cite:CVPR15DNNH} improved the two-stage CNNH with a simultaneous feature learning and hash coding  pipeline such that representations and hash codes can be optimized in a joint learning process. DHN \cite{cite:AAAI16DHN} further improves DNNH by a cross-entropy loss and a quantization loss which preserve the pairwise similarity and control the quantization error simultaneously. HashNet \cite{cite:ICCV17HashNet} attack the ill-posed gradient problem of sign by continuation, which directly optimized the sign function. HashNet obtains state-of-the-art performance on several benchmarks.

However, prior hash methods perform not so good within Hamming Radius 2 since their loss penalize little on small Hamming distance. And they suffer from large distribution gap between domains under the transfer setting. DVSH~\cite{cite:KDD16DVSH} and PRDH~\cite{cite:AAAI17PRDH} integrate different types of pairwise constraints to encourage the similarities of the hash codes from an intra-modal view and an inter-modal view, with additional decorrelation constraints for enhancing the discriminative ability of each hash bit. 
THN~\cite{cite:AAAI17THN} aligns the distribution of database domain with auxiliary domain by minimize the Maximum Mean Discrepancy (MMD) of hash codes in Hamming Space, which fits the transfer setting. 

However, adversarial learning has been applied to transfer learning~\cite{cite:ICML15RevGrad} and achieves the state of the art performance. Thus, the proposed Transfer Adversarial Hashing addresses distribution gap between source and target domain by adversarial learning. With similarity relationship learning designed for searching in Hamming Radius 2 and adversarial learning for domain distribution alignment, TAH can solve the transfer setting for image retrieval efficiently and effectively.

\begin{figure*}[tbp]
  \centering
  \includegraphics[width=0.8\textwidth]{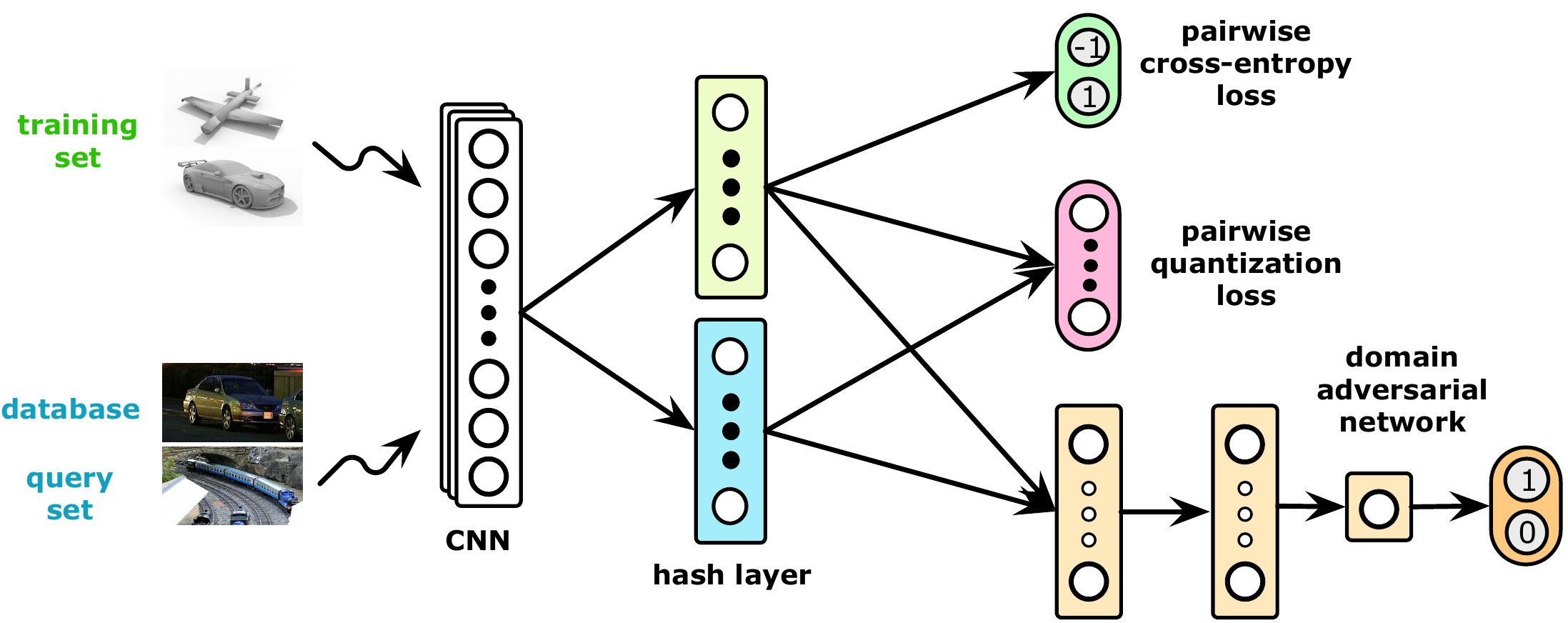}
  \vspace{-10pt}
  \caption{The architecture of transfer adversarial hashing (TAH), which consists of three modules: similarity relationship learning (green), domain distribution alignment (orange), and quantization error minimization (pink).}
   \label{fig:TAH}
   \vspace{-10pt}
\end{figure*}

\section{Transfer Adversarial Hashing}
In transfer retrieval setting, we are given a database ${\cal Y} = \{ {\bm y}_k\}_{k=1}^m$ from target domain $Y$ and a training set ${\cal X}  = \{ {\bm x}_i\}_{i=1}^{n}$ from source domain $X$, where ${\bm x}_i,{\bm y}_k \in \mathcal{\bm{R}}^{d}$ are $d$-dimensional feature vectors. The key challenge of transfer hashing is that no supervised relationship is available between database points. Hence, we build a hashing model for the database of target domain $Y$ by learning from a training dataset ${\cal X}  = \{ {\bm x}_i\}_{i=1}^{n}$ available in a different but related source domain $X$, which consists of similarity relationship $\mathcal{S} = \{ s_{ij}\}$, where $s_{ij} = 1$ implies points ${{\bm x}}_i$ and ${{\bm x}}_j$ are similar while $s_{ij} = 0$ indicates points ${{\bm x}}_i$ and ${{\bm x}}_j$ are dissimilar. In real image retrieval applications, the similarity relationship ${\cal S}=\{s_{ij}\}$ can be constructed from the semantic labels among the data points or the relevance feedback from click-through data in online image retrieval systems.

The goal of Transfer Adversarial Hashing (TAH) is to learn a hash function $f :\mathbb{R}^{d} \to \{ -1,1\}^b$ encoding data points $x$ and $y$ from domains $X$ and $Y$ into compact $b$-bit hash codes ${\bm h^x} = f({\bm x})$ and ${\bm h^y} = f({\bm y})$, such that both ground truth similarity relationship ${\cal S}$ for domain $X$ and the unknown similarity relationship ${\cal S'}$ for domain $Y$ can be preserved. With the learned hash function, we can generate hash codes ${\cal H}^x=\{{\bm h}_i^x\}_{i=1}^n$ and ${\cal H}^y=\{{\bm h}_k^y\}_{k=1}^m$ for the training set and database respectively, which enables image retrieval in the Hamming space through ranking the Hamming distances between hash codes of the query and database points.

\subsection{The Overall Architecture}
The architecture for learning the transfer hash function is shown in Figure~\ref{fig:TAH}, which is a hybrid deep architecture of a deep hashing network and a domain adversarial network. In the deep hashing network $G_f$, we extend AlexNet \cite{cite:NIPS12CNN}, a deep convolutional neural network (CNN) comprised of five convolutional layers $conv1$--$conv5$ and three fully connected layers $fc6$--$fc8$. We replace the $fc8$ layer with a new $fch$ hash layer with $b$ hidden units, which transforms the network activation ${\bm z}_i^x$ in $b$-bit hash code by sign thresholding ${\bm h}^x_{i} = \operatorname{sgn} ({\bm z}_i^{x})$. Since it is hard to optimize sign function for its ill-posed gradient, we adopt the hyperbolic tangent (tanh) function to squash the activations to be within $[-1,1]$, which reduces the gap between the $fch$-layer representation ${\bm z}_i^{\ast}$ and the binary hash codes ${\bm h}_i^{\ast}$, where $\ast\in\{x,y\}$. And a pairwise $t$-distribution cross-entropy loss and a pairwise quantization loss are imposed on the hash codes. In domain adversarial network $G_d$, we use the Multilayer Perceptrons (MLP) architecture adopted by~\cite{cite:ICML15RevGrad}. It accepts as inputs the hash codes generated by the deep hashing network $G_f$ and consists of three fully connected layers, with the numbers of units being $(b,1024,1024,1)$. 
The last layer of $G_d$ output the probability of the input data belonging to a specific domain. And a cross-entropy loss is added on the output of the adversarial network. This hybrid deep network can achieve hash function learning through similarity relationship preservation and domain distribution alignment simultaneously, which enables image retrieval from the database in  the target domain.

\subsection{Hash Function Learning}

To perform deep learning to hash from image data, we jointly preserve similarity relationship information underlying pairwise images and generate binary hash codes by Maximum A Posterior (MAP) estimation.

\begin{figure}[!tbp]
    \centering
    \subfigure[Probability]{
        \includegraphics[width=0.22\textwidth]{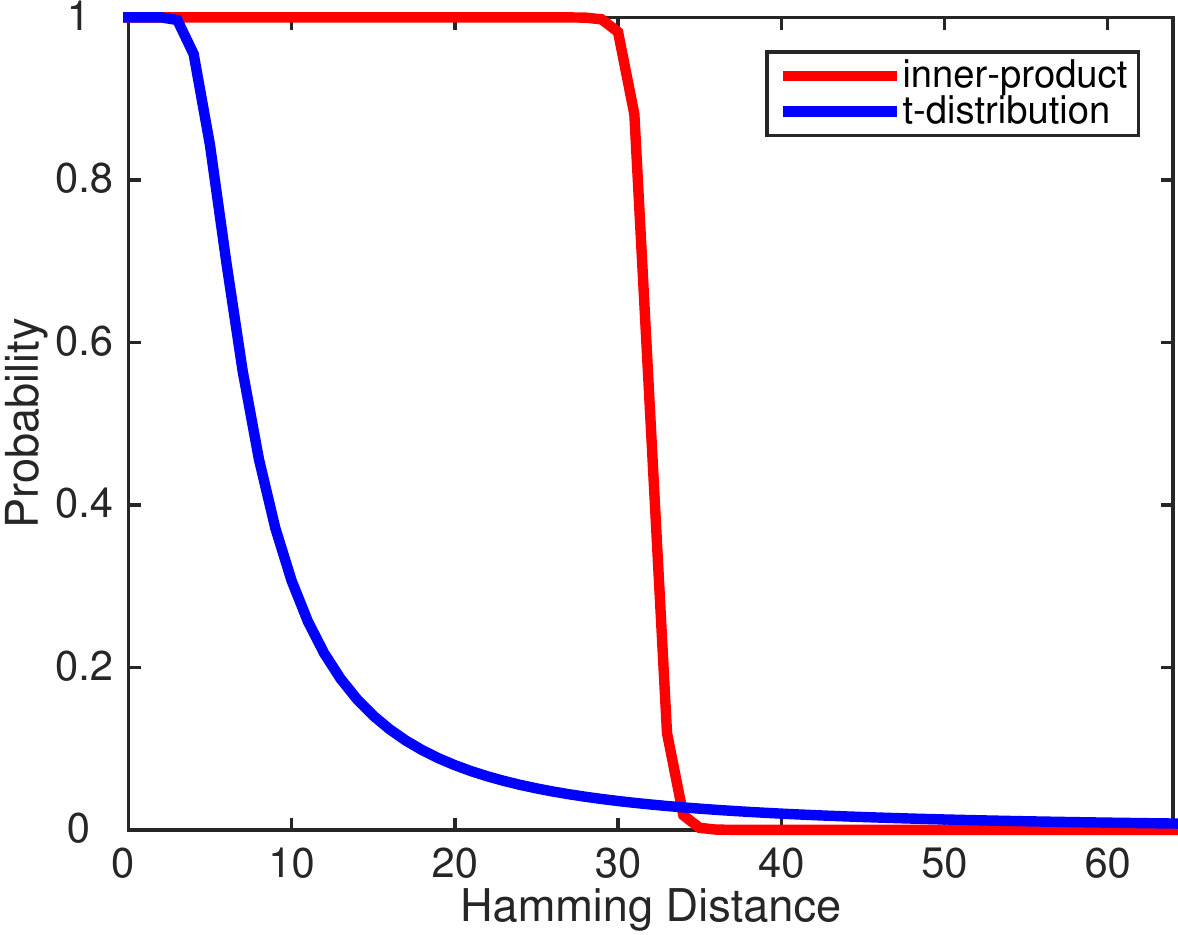}
        \label{fig:formula_probability}
    }
    \subfigure[Loss Value]{
        \includegraphics[width=0.22\textwidth]{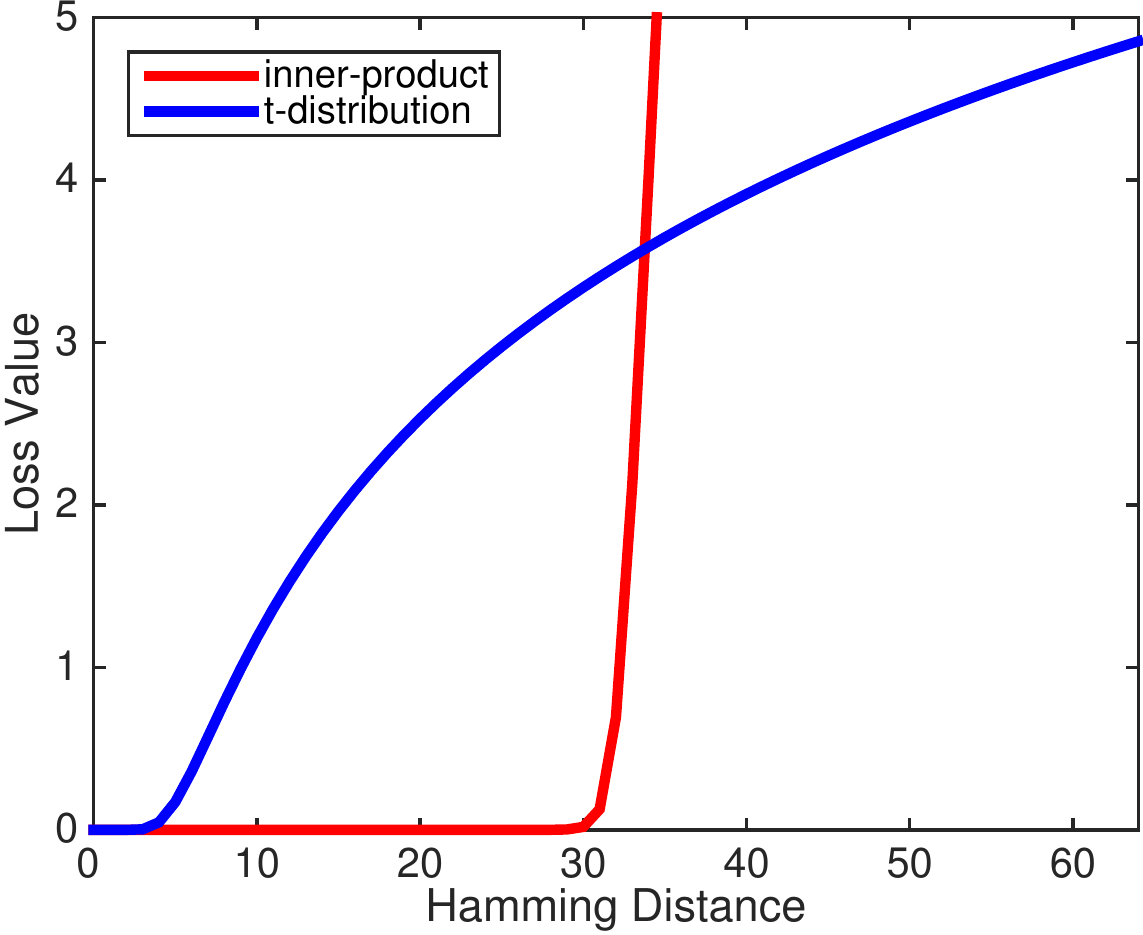}
        \label{fig:formula_loss}
    }
    \caption{Probability (a) and Loss value (b) w.r.t Hamming Distance between codes for similar data points.}
    \label{fig:formula}
\end{figure}

Given the set of pairwise similarity labels $\mathcal{S} = \{s_{ij}\}$, the logarithm Maximum a Posteriori (MAP) estimation of training hash codes ${\bm H}^x = [{\bm h}_1^x,\ldots,{\bm h}_{n}^x]$ can be defined as
\begin{equation}\label{eqn:MAP}
	\small
	\begin{aligned}
  	\log p\left( {{\bm{H}^x}|\mathcal{S}} \right) &\propto \log p\left( {\mathcal{S}|{\bm{H}^x}} \right) p\left( {{{\bm{H}}^x}} \right) \\
    &= \sum\limits_{{s_{ij}} \in \mathcal{S}} {\log p\left( {{s_{ij}}|{{\bm{h}}_i^x},{{\bm{h}}_j^x}} \right)p\left( {{{\bm{h}}_i^x}} \right)p\left( {{{\bm{h}}_j^x}} \right)} ,
	\end{aligned}
	\normalsize
\end{equation}
where  $p({\cal S}|{\bm H}^x)$ is likelihood function, and $p({{\bm H}}^x)$ is prior distribution. For each pair of points ${\bm x}_i$ and ${\bm x}_j$, $p(s_{ij}|{\bm h}_i^x,{\bm h}_j^x)$ is the conditional probability of their relationship $s_{ij}$ given their hash codes ${\bm h}_i^x$ and ${\bm h}_j^x$, which can be defined using the pairwise logistic function,
 \begin{equation}\label{eqn:CDF}
	\begin{aligned}
	  & p \left( {{s_{ij}}|{{\bm h}_i^x},{{\bm h}_j^x}} \right) = 
		\begin{cases}
			\sigma \left( \text{sim} \left({\bm h}_i^x, {\bm h}_j^x \right) \right), & {s_{ij}} = 1 \\
			1 - \sigma \left( \text{sim} \left({\bm h}_i^x, {\bm h}_j^x \right) \right), & {s_{ij}} = 0 \\
		\end{cases} \\
	    & = \sigma {\left( \text{sim} \left({\bm h}_i^x, {\bm h}_j^x \right) \right)^{{s_{ij}}}}{\left( {1 - \sigma \left( \text{sim} \left({\bm h}_i^x, {\bm h}_j^x \right) \right)} \right)^{1 - {s_{ij}}}},
	\end{aligned}
	\normalsize
\end{equation}
where $\text{sim} \left({\bm h}_i^x, {\bm h}_j^x \right)$ is the similarity function of code pairs ${\bm h}_i^x$ and ${\bm h}_j^x$ and $\sigma \left( x \right)$ is the probability function. Previous methods~\cite{cite:AAAI16DHN,cite:ICCV17HashNet} usually adopt inner product function $\left\langle {\bm h}_i^x, {\bm h}_j^x \right\rangle $ as similarity function and $\sigma \left( x \right) = {1}/({{1 + {e^{ - \alpha x}}}})$ as probability function. However, from Figure~\ref{fig:formula}, we can observe that the probability corresponds to these similarity function and probability function stays high when the Hamming distance between codes is larger than 2 and only starts to decrease when the Hamming distance becomes close to $b/2$ where $b$ is the number of hash bits. This means that previous methods cannot force the Hamming distance between codes of similar data points to be smaller than 2 since the probability cannot discriminate different Hamming distances smaller than $b/2$ sufficiently.

To tackle the above mis-specification of the inner product, we proposes a new similarity function inspiring by the success of $t$-distribution with one degree of freedom for modeling long-tail dataset,
\begin{equation}
	\text{sim} \left({\bm h}_i^x, {\bm h}_j^x \right) = \frac{b}{1+\left\lVert {{\bm h}_i^x-{\bm h}_j^x} \right\rVert^2 }, 
\end{equation}
and the corresponding probability function is defined as $\sigma \left( x \right) = \tanh \left(\alpha x \right)$. Similar to previous methods, these functions also satisfy that the smaller the Hamming distance ${\text{dis}}{{\textrm{t}}_H}( {{{\bm{h}}_i^x},{{\bm{h}}_j^x}} )$ is, the larger the similarity function value  $\text{sim} \left({\bm h}_i^x, {\bm h}_j^x \right)$ will be, and the larger $p( {1|{{\bm{h}}_i^x},{{\bm{h}}_j^x}} )$ will be, implying that pair ${\bm h}_i^x$ and ${\bm h}_j^x$ should be classified as ``similar''; otherwise, the larger $p( {0|{{\bm{h}}_i^x},{{\bm{h}}_j^x}} )$ will be, implying that pair ${\bm h}_i^x$ and ${\bm h}_j^x$ should be classified as ``dissimilar''. Furthermore, from Figure~\ref{fig:formula}, we can observe that our probability w.r.t Hamming distance between code pairs decreases significantly when the Hamming distance is larger that $2$, indicating that our loss function will penalize Hamming distance larger than $2$ for similar codes much more than previous methods. Thus, our similarity function and probability function perform better for search within Hamming Radius $2$. Hence, Equation~\eqref{eqn:CDF} is a reasonable extension of the logistic regression classifier which optimizes the performance of searching within Hamming Radius 2 of a query.

Similar to previous work~\cite{cite:AAAI14CNNH,cite:CVPR15DNNH,cite:AAAI16DHN}, defining that ${\bm h}_i^x = \operatorname{sgn}({{\bm z}_i^x})$ where ${\bm z}_i^x$ is the activation of hash layer, we relax binary codes to continuous codes since discrete optimization of Equation~\eqref{eqn:MAP} with binary constraints is difficult and adopt a quantization loss function to control quantization error. Specifically, we adopt the prior for quantization of~\cite{cite:AAAI16DHN} as 
\begin{equation}
p\left( {{\bm z}_i^x} \right) = \frac{1}{{2\varepsilon }}\exp \left( { - \frac{\|{|{\bm z_i^x}| - {\bm 1}}\|_1}{\varepsilon }} \right)\
\label{eqn:prior}
\end{equation}
where $\varepsilon$ is the parameter of the exponential distribution.

By substituting Equations \eqref{eqn:CDF} and \eqref{eqn:prior} into the MAP estimation in Equation~\eqref{eqn:MAP}, we achieve the optimization problem for similarity hash function learning as follows,
\begin{equation}\label{eqn:HRL}
\mathop {\min }\limits_\Theta  J = L + \lambda Q, \\
\end{equation}
where $\lambda$ is the trade-off parameter between pairwise cross-entropy loss $L$ and pairwise quantization loss $Q$, and $\Theta$ is a set of network parameters. Specifically, loss $L$ is defined as
\begin{equation}\label{eqn:heteoL}
\begin{aligned}
  {L} & = \sum\limits_{s_{ij}\in{\cal S}} \log \left( \frac{2}{1+\exp \left(2\alpha \frac{b}{1+\left\lVert {{\bm z}_i^x-{\bm z}_j^x} \right\rVert_2 } \right)} \right) \\
  &+ {s_{ij}} \log \left( \frac{\exp \left(2\alpha \frac{b}{1+\left\lVert {{\bm z}_i^x-{\bm z}_j^x} \right\rVert_2 } \right) -1}{2} \right).
\end{aligned}
\end{equation}
Similarly the pairwise quantization loss $Q$ can be derived as
\begin{equation}\label{eqn:Q}
	Q = \sum\limits_{{s_{ij}} \in \mathcal{S}} {\left( {{{\left\| {\left| {{\bm{z}}_i^x} \right| - {\bm{1}}} \right\|}_1} + {{\left\| {\left| {{\bm{z}}_j^x} \right| - {\bm{1}}} \right\|}_1}} \right)} ,
\end{equation}
where ${\bm 1} \in \mathbb{R}^K$ is the vector of ones.
By the MAP estimation in Equation \eqref{eqn:HRL}, we can simultaneously preserve the similarity relationship and control the quantization error of binarizing continuous activations to binary codes in source domain.

\subsection{Domain Distribution Alignment}
The goal of transfer hashing is to train the model on data of source domain and perform efficient retrieval from the database of target domain in response to the query of target domain. Since there is no relationship between the  database points, we exploit the training data ${{\cal X}}$ to learn the relationship among the database points. However, there is large distribution gap between the source domain and the target domain. Therefore, we should further reduce the distribution gap between the source domain and the target domain in the Hamming space. 

Domain adversarial networks have been successfully applied to transfer learning~\cite{cite:ICML15RevGrad} by extracting transferable features that can reduce the distribution shift between the source domain and the target domain. Therefore, in this paper, we reduce the distribution shifts between the source domain and the target domain by adversarial learning. The adversarial learning procedure is a two-player game, where the first player is the domain discriminator $G_d$ trained to distinguish the source domain from the target domain, and the second player is the base hashing network $G_f$ fine-tuned simultaneously to confuse the domain discriminator.

To extract domain-invariant hash codes $\bm{h}$, the parameters $\theta_f$ of deep hashing network $G_f$ are learned by maximizing the loss of domain discriminator $G_d$, while the parameters $\theta_d$ of domain discriminator $G_d$ are learned by minimizing the loss of the domain discriminator. The objective of domain adversarial network is the functional:
\begin{equation}\label{eqn:GRL}
	D \left( {{\theta _f},{\theta _d}} \right) = \frac{1}{{{n} + {m}}}\sum\limits_{{{\bm{u}}_i} \in {{\cal X} \cup {\cal Y}}} {{L_d}\left( {{G_d}\left( {{G_f}\left( {{{\bm{u}}_i}} \right)} \right),{d_i}} \right)} ,
\end{equation}
where $L_d$ is the cross-entropy loss and $d_i$ is the domain label of data point $\bm{u}_i$. $d_i = 1$ means $\bm{u}_i$ belongs to target domain and $d_i = 0$ means $\bm{u}_i$ belongs to source domain.
Thus, we define the overall loss by integrating Equations \eqref{eqn:HRL} and \eqref{eqn:GRL},
\begin{equation}\label{eqn:model}
   C = J - \mu D,
\end{equation}
where $\mu$ is a trade-off parameter between the MAP loss $J$ and adversarial learning loss $D$. The optimization of this loss is as follows. After training convergence, the parameters $\hat\theta_f$ and $\hat\theta_d$ will deliver a saddle point of the functional~\eqref{eqn:model}: 
\begin{equation}\label{eqn:param1}
\begin{gathered}
     (\hat\theta_f) = \arg \mathop {\min }\limits_{{\theta _f}} C \left( {{\theta _f},{\theta _d}} \right), \\
     (\hat\theta_d) = \arg \mathop {\max }\limits_{{\theta_d}} C \left( {{\theta _f},{\theta _d}} \right).
\end{gathered}
\end{equation}
This mini-max problem can be trained end-to-end by back-propagation over all network branches in Figure~\ref{fig:TAH}, where the gradient of the adversarial loss $D$ is reversed and added to the gradient of the hashing loss $J$.
By optimizing the objective function in Equation~\eqref{eqn:model}, we can learn transfer hash codes which preserve the similarity relationship and align the domain distributions as well as control the quantization error of sign thresholding. Finally, we generate $b$-bit hash codes by sign thresholding as ${\bm h}^\ast = {\mathop{\rm sgn}} (\bm{z}^\ast)$, where  ${\mathop{\rm sgn}} (\bm{z})$ is the sign function on vectors that for each dimension $i$ of $\bm{z}^\ast$, $k=1,2,...,b$, ${\mathop{\rm sgn}} (z_k^\ast)=1$ if $z_k^\ast > 0$, otherwise ${\mathop{\rm sgn}} (z_k^\ast)=-1$. Since the quantization error in Equation \eqref{eqn:model} has been minimized, this final binarization step will incur small loss of retrieval quality for transfer hashing.

\section{Experiments}\label{section:Experiments}
We extensively evaluate the efficacy of the proposed TAH model against state of the art hashing methods on two benchmark datasets. The codes and configurations will be made available online.

\subsection{Setup}

\textbf{NUS-WIDE}\footnote{\scriptsize{\url{http://lms.comp.nus.edu.sg/research/NUS-WIDE.htm}}} is a popular dataset for cross-modal retrieval, which contains 269,648 image-text pairs. The annotation for 81 semantic categories is provided for evaluation. We follow
the settings in \cite{cite:AAAI16DHN,cite:ICML11AGH,cite:CVPR15DNNH} and use the
subset of 195,834 images that are associated with the 21
most frequent concepts, where each concept consists of at
least 5,000 images. Each image is resized into $256 \times 256$ pixels. We follow similar experimental protocols as DHN \cite{cite:AAAI16DHN} and randomly sample 100 images per category as queries, with the remaining images used as the database; furthermore, we randomly sample 500 images per category (each image attached to one category in sampling) from the database as training points.

\textbf{VisDA2017}\footnote{\scriptsize{\url{https://github.com/VisionLearningGroup/taskcv-2017-public/tree/master/classification}}} is a cross-domain image dataset of images rendered from CAD models as synthetic image domain and real object images cropped from the COCO dataset as real image domain. We perform two types of transfer retrieval tasks on the VisDA2017 dataset: (1) using real image query to retrieve real images where the training set consists of synthetic images  (denoted by $synthetic \rightarrow real$); (2) using synthetic image query to retrieve synthetic images where the training set consists of real images (denoted by $real \rightarrow synthetic$). The relationship ${\cal S}$ for training and the ground-truth for evaluation are defined as follows: if an image $i$ and a image $j$ share the same category, they are relevant, i.e. $s_{ij}=1$; otherwise, they are irrelevant, i.e. $s_{ij}=0$. Similarly, we randomly sample 100 images per category of target domain as queries, and use the remaining images of target domain as the database and we randomly sample 500 images per category from both source domain and target domain as training points, where source domain data points have ground truth similarity information while the target domain data points do not.

We use retrieval metrics within Hamming radius 2 to test the efficacy of different methods. We evaluate the retrieval quality based on standard evaluation metrics: Mean Average Precision (MAP), Precision-Recall curves and Precision all within Hamming radius 2. We compare the retrieval quality of our \textbf{TAH} with ten classical or state-of-the-art hashing methods, including unsupervised methods \textbf{LSH} \cite{cite:VLDB99LSH}, \textbf{SH} \cite{cite:NIPS09SH}, \textbf{ITQ} \cite{cite:CVPR11ITQ}, supervised shallow methods \textbf{KSH} \cite{cite:CVPR12KSH}, \textbf{SDH} \cite{cite:CVPR15SDH}, supervised deep single domain methods  \textbf{CNNH} \cite{cite:AAAI14CNNH}, \textbf{DNNH} \cite{cite:CVPR15DNNH}, \textbf{DHN} \cite{cite:AAAI16DHN}, \textbf{HashNet} \cite{cite:ICCV17HashNet} and supervised deep cross-domain method \textbf{THN} \cite{cite:AAAI17THN}.

For fair comparison, all of the methods use identical training and test sets. For deep learning based methods, we directly use the image pixels as input. For the shallow learning based methods, we reduce the 4096-dimensional AlexNet features \cite{cite:ICML14DeCAF} of images. We adopt the AlexNet architecture \cite{cite:NIPS12CNN} for all deep hashing methods, and implement TAH based on the \textbf{Caffe} framework \cite{cite:MM14Caffe}. For the single domain task on NUS-WIDE, we test cross-domain method TAH and THN by removing the transfer part. For the cross-domain tasks on VisDA2017, we train single domain methods with data of source domain and directly apply the trained model to the query and database of another domain. We fine-tune convolutional layers $conv1$--$conv5$ and fully-connected layers $fc6$--$fc7$ copied from the AlexNet model pre-trained on ImageNet 2012 and train the hash layer $fch$ and adversarial layers, all through back-propagation. As the $fch$ layer and the adversarial layers are trained from scratch, we set its learning rate to be 10 times that of the lower layers. We use mini-batch stochastic gradient descent (SGD) with 0.9 momentum and the learning rate annealing strategy implemented in Caffe. The penalty of adversarial networks $mu$ is increased from 0 to 1 gradually as RevGrad~\cite{cite:ICML15RevGrad}. We cross-validate the learning rate from $10^{-5}$ to $10^{-3}$ with a multiplicative step-size ${10}^{\frac{1}{2}}$. We fix the mini-batch size of images as $64$ and the weight decay parameter as $0.0005$.

\begin{table*}[tb]
    \centering 
    \caption{MAP Results of Ranking within Hamming Radius 2 for Different Number of Bits on Three Image Retrieval Tasks}
    \label{table:MAP}
	  \resizebox{\textwidth}{!}{
    \begin{tabular}{c|cccc|cccc|cccc}
        \Xhline{1.0pt}
        \multirow{3}{30pt}{\centering Method} & 
        \multicolumn{4}{c|}{\multirow{2}{50pt}{\centering NUS-WIDE}} & \multicolumn{8}{c}{VisDA2017}\\
        \cline{6-13}
        & \multicolumn{4}{c|}{} & \multicolumn{4}{c|}{$synthetic \rightarrow real$} & \multicolumn{4}{c}{$real \rightarrow synthetic$} \\
        \cline{2-13}
        & 16 bits & 32 bits  & 48 bits  & 64 bits  & 16 bits & 32 bits  & 48 bits  & 64 bits  & 16 bits & 32 bits  & 48 bits  & 64 bits \\
        \hline
        TAH & \textbf{0.722} & \textbf{0.729} & \textbf{0.692} & \textbf{0.680} & \textbf{0.465} & \textbf{0.423}  & \textbf{0.433} & \textbf{0.404} & \textbf{0.672} & \textbf{0.695} & \textbf{0.784} & \textbf{0.761} \\
        THN & 0.671 & 0.676 & 0.662 & \underline{0.603} & \underline{0.415} & 0.396 & 0.228 & 0.127 & \underline{0.647} & \underline{0.687} & \underline{0.664} & 0.532 \\
        HashNet & \underline{0.709} & \underline{0.693} & \underline{0.681} & 0.615 & 0.412 & \underline{0.403} & \underline{0.345} & 0.274 & 0.572 & 0.676 & 0.662 & \underline{0.642} \\
        DHN & 0.669 & 0.672 & 0.661 & 0.598 & 0.331 & 0.354 & 0.309 & \underline{0.281} & 0.545 & 0.612 & 0.608 & 0.604 \\
        DNNH & 0.568 & 0.622 & 0.611 & 0.585 & 0.241 & 0.276 & 0.252 & 0.243 & 0.509 & 0.564 & 0.551 & 0.503 \\
        CNNH & 0.542 & 0.601 & 0.587 & 0.535 & 0.221 & 0.254 & 0.238 & 0.230 & 0.487 & 0.568 & 0.530 & 0.445 \\
        SDH & 0.555 & 0.571 & 0.517 & 0.499 & 0.196 & 0.238 & 0.229 & 0.212 & 0.330 & 0.388 & 0.339 & 0.277 \\
        ITQ & 0.498 & 0.549 & 0.517 & 0.402 & 0.187 & 0.175 & 0.146 & 0.123 & 0.163 & 0.193 & 0.176 & 0.158 \\
        SH & 0.496 & 0.543 & 0.437 & 0.371 & 0.154 & 0.141 & 0.130 & 0.105 & 0.154 & 0.182 & 0.145 & 0.123 \\
        KSH & 0.531 & 0.554 & 0.421 & 0.335 & 0.176 & 0.183 & 0.124 & 0.085 & 0.143 & 0.178 & 0.146 & 0.092 \\
        LSH & 0.432 & 0.453 & 0.323 & 0.255 & 0.122 & 0.092 & 0.083 & 0.071 & 0.130 & 0.145 & 0.122 & 0.063 \\
        \Xhline{1.0pt}
    \end{tabular}
    }
\end{table*}

\begin{figure*}[!phtb]
    \centering
    \subfigure[NUS-WIDE]{
        \includegraphics[width=0.28\textwidth]{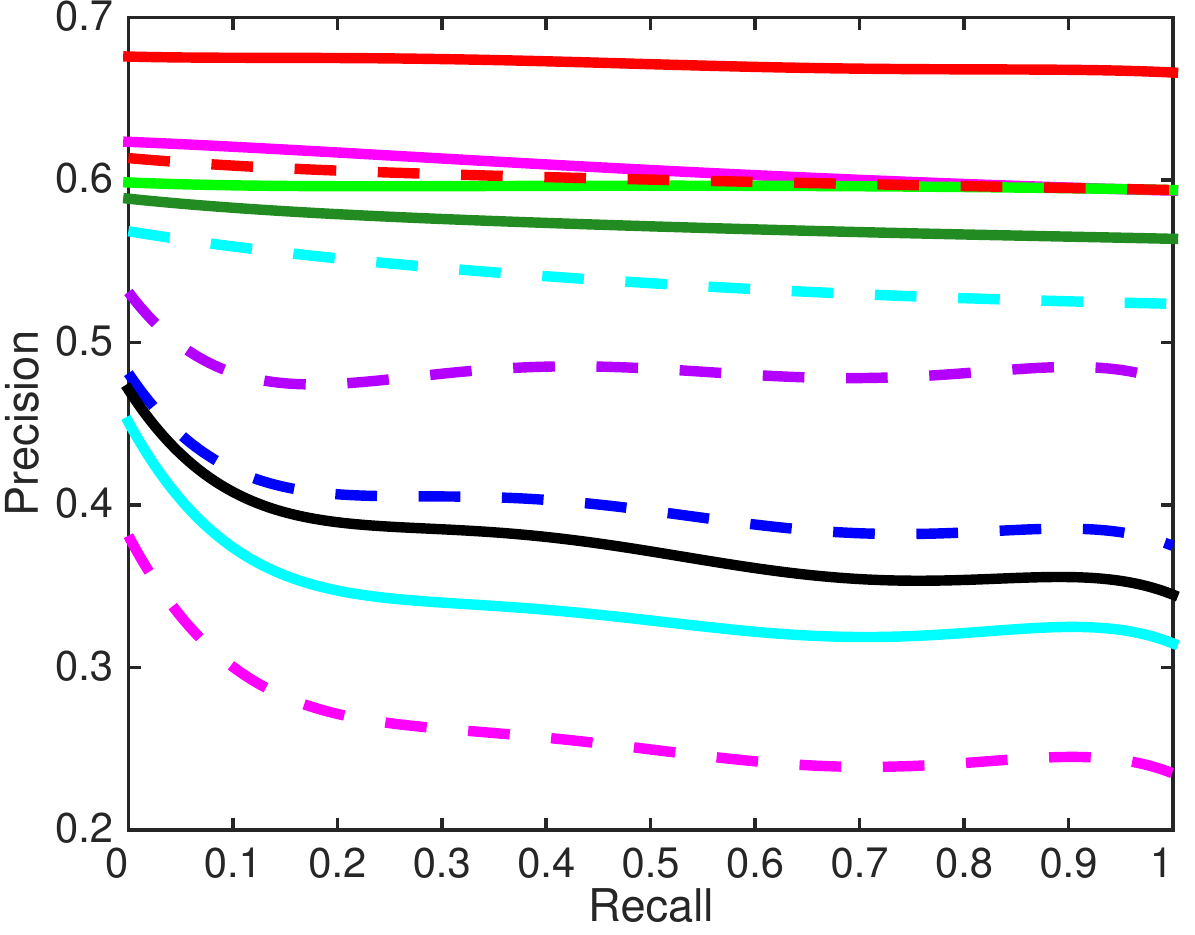}
        \label{fig:pr_nus}
    }
		\hfil
    \subfigure[synthetic $\rightarrow$ real]{
        \includegraphics[width=0.28\textwidth]{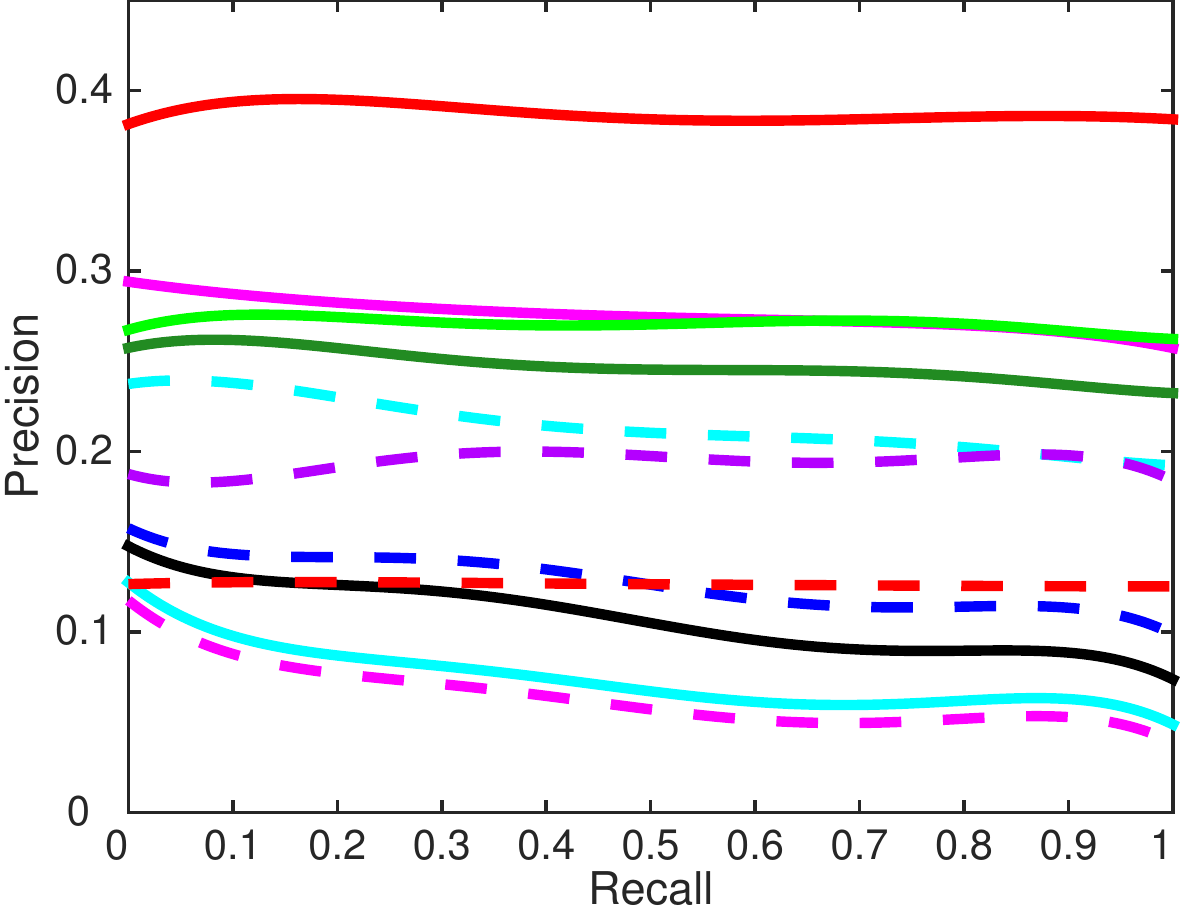}
        \label{fig:pr_train}
    }
		\hfil
    \subfigure[real $\rightarrow$ synthetic]{
        \includegraphics[width=0.36\textwidth]{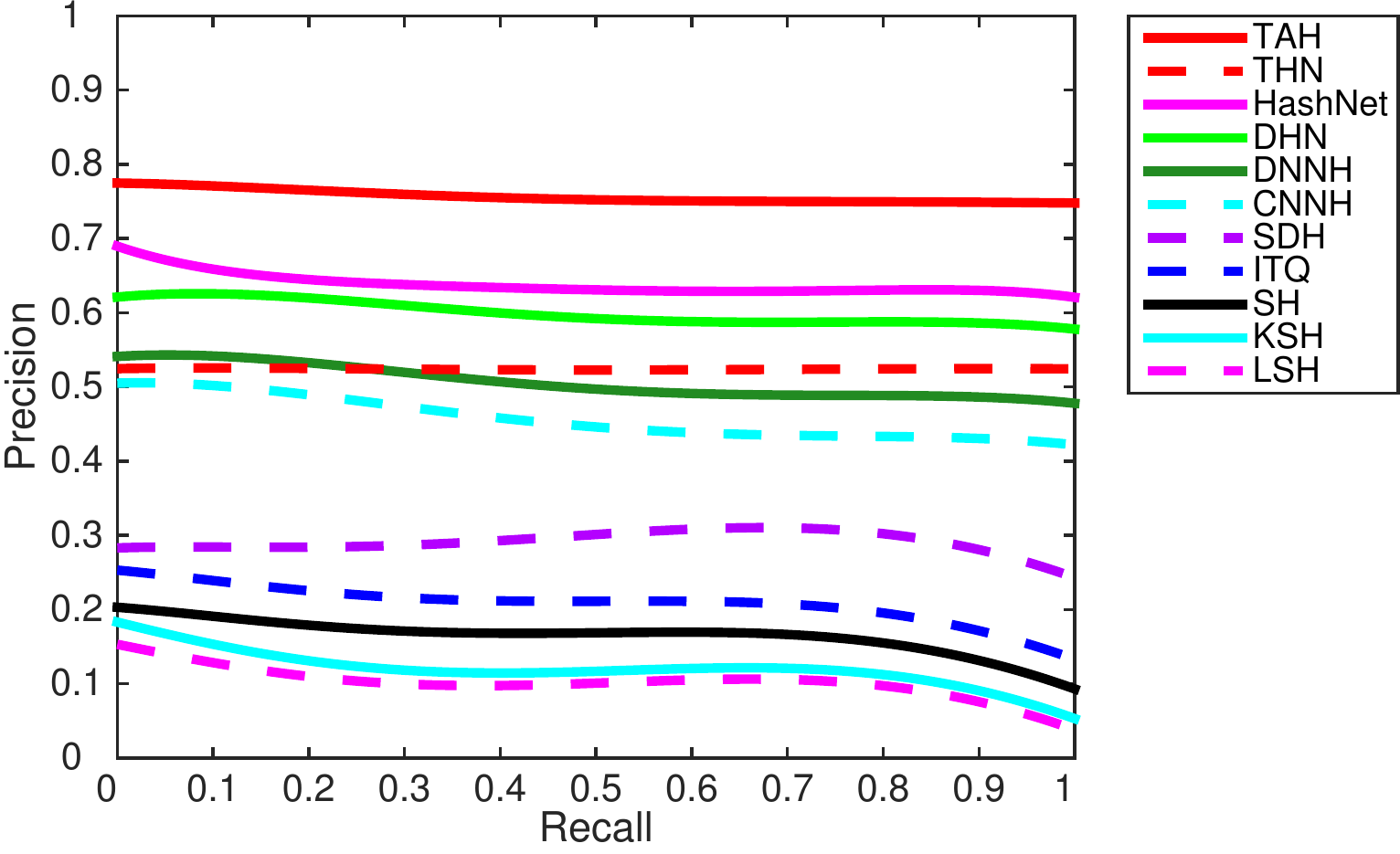}
        \label{fig:pr_validation}
    }
    \vspace{-10pt}
    \caption{The Precision-recall curve @ 64 bits within Hamming radius 2 of TAH and comparison methods on three tasks.}
    \label{fig:pr}
\end{figure*}

\begin{figure*}[!phtb]
    \centering
    \subfigure[NUS-WIDE]{
        \includegraphics[width=0.28\textwidth]{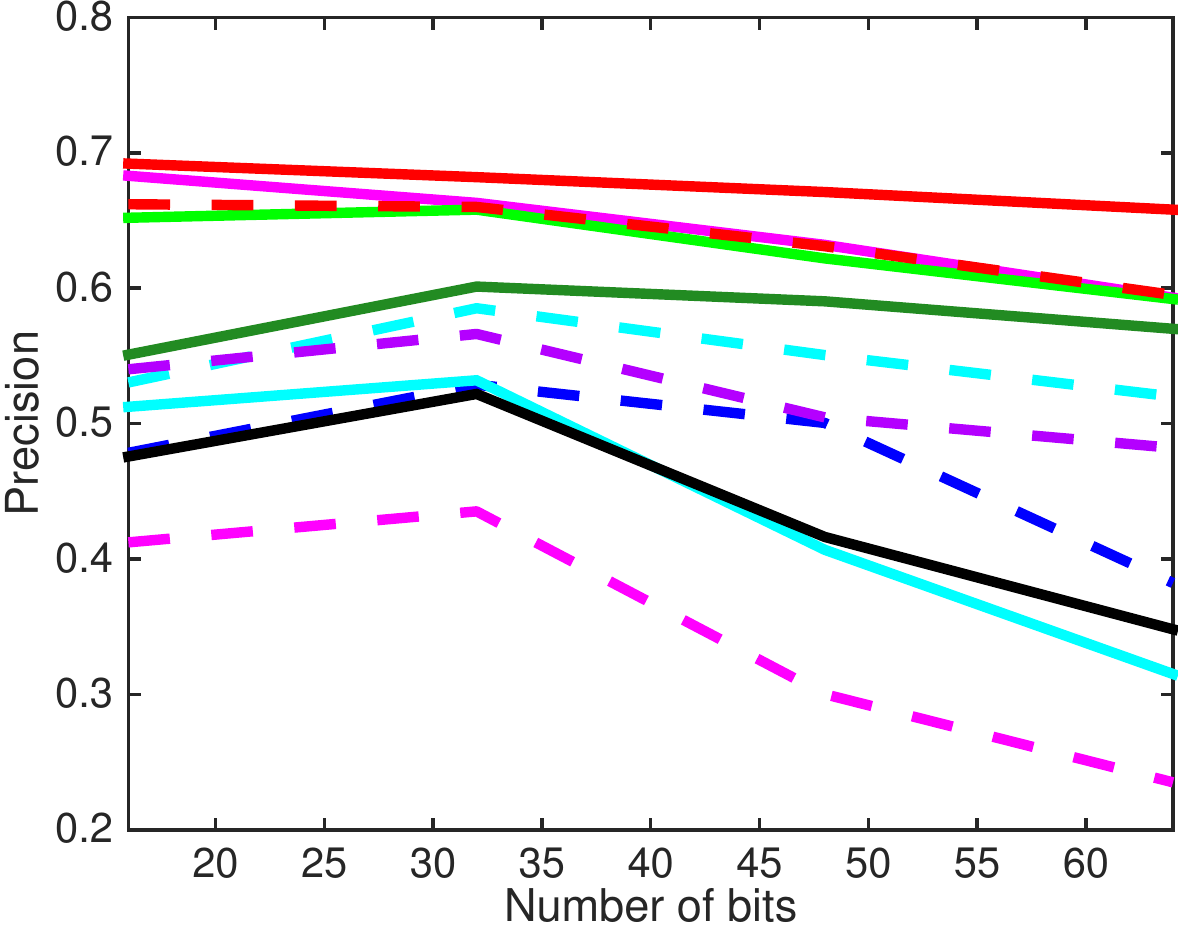}
        \label{fig:ham_nus}
    }
		\hfil
    \subfigure[synthetic $\rightarrow$ real]{
        \includegraphics[width=0.28\textwidth]{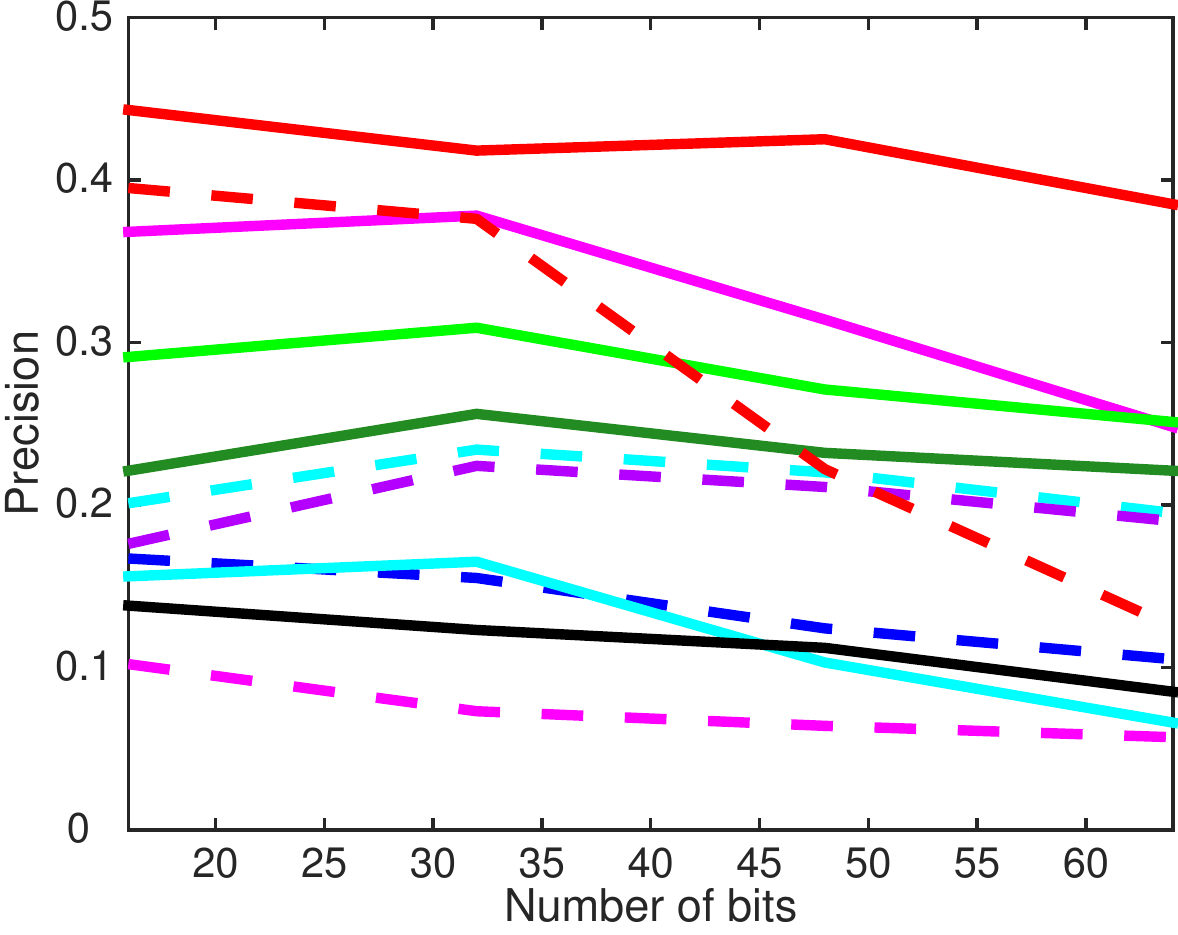}
        \label{fig:ham_train}
    }
		\hfil
    \subfigure[real $\rightarrow$ synthetic]{
        \includegraphics[width=0.36\textwidth]{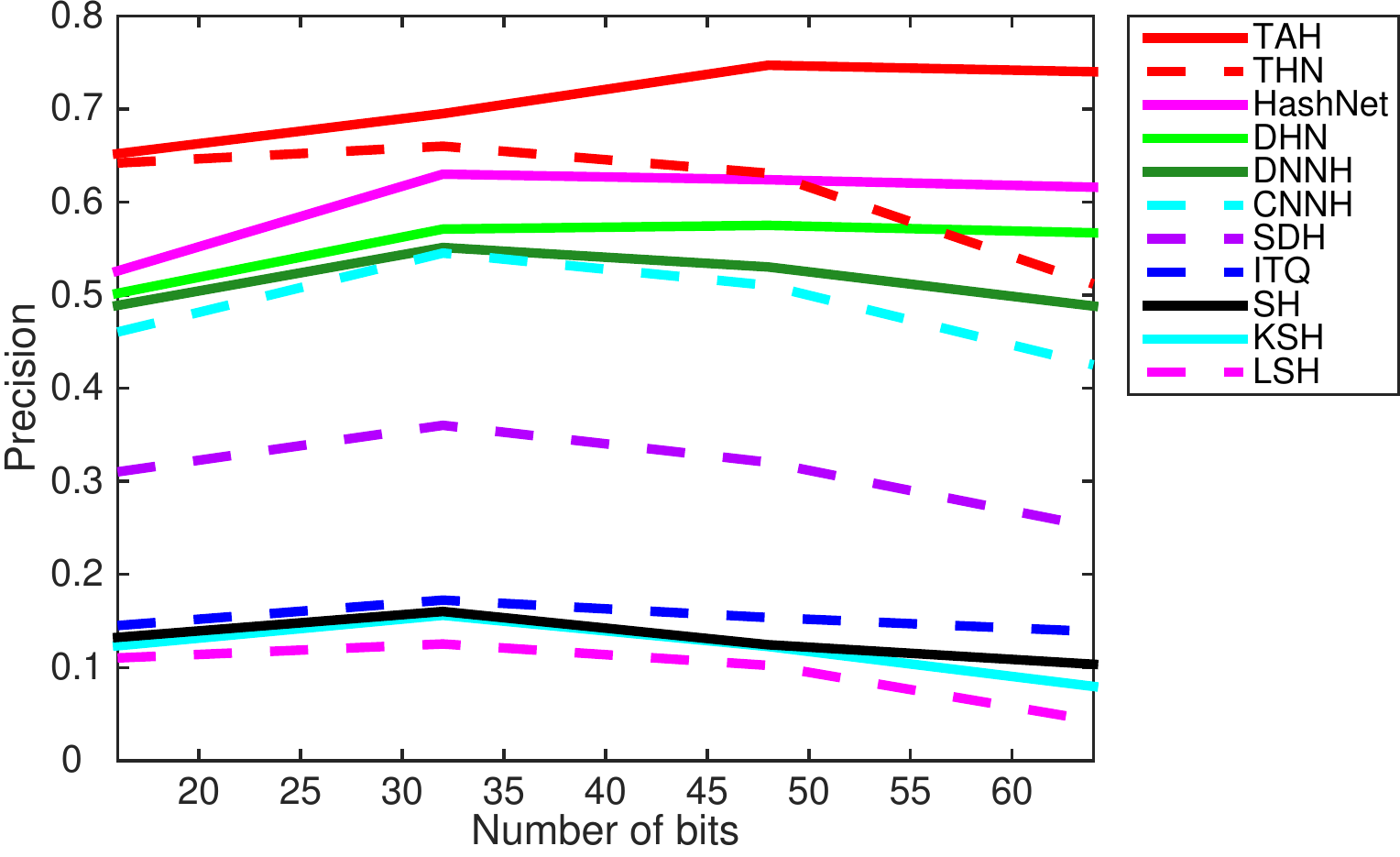}
        \label{fig:ham_validation}
    }
    \vspace{-10pt}
    \caption{The Precision curves @64 bits within Hamming radius 2 of TAH and comparison methods on three tasks.}
    \label{fig:ham}
\end{figure*}

\subsection{Results}
\textbf{NUS-WIDE:} The Mean Average Precision (MAP) within Hamming Radius 2 results are shown in Table \ref{table:MAP}. We can observe that on the classical task that database and query images are from the same domain, TAH generally outperforms state of the art methods defined on classical retrieval setting. Specifically, compared to the best method on this task, HashNet, and state of the art cross-domain method THN, we achieve absolute boosts of 0.031 and 0.053 in average MAP for different bits on NUS-WIDE, which is very promising. 

The precision-recall curves within Hamming Radius 2 based on 64-bits hash codes for the NUS-WIDE dataset are illustrated in Figure~\ref{fig:pr_nus}. We can observe that TAH achieves the highest precision at all recall levels. The precision nearly does not decrease with the increasing of recall, proving that TAH has stable performance for Hamming Radius 2 search.

The Precision within Hamming radius 2 curves are shown in Figure~\ref{fig:ham_nus}. We can observe that TAH achieves the highest P@H=2 results on this task. When using longer codes, the Hamming space will become sparse and few data points fall within the Hamming ball with radius 2 \cite{cite:CVPR12MIH}. This is why most hashing methods perform worse on accuracy with very long codes. However, TAH achieves a relatively mild decrease on accuracy with the code length increasing. This validates that TAH can concentrate hash codes of similar data points to be within the Hamming ball of radius $2$.

These results validate that TAH is robust under diverse retrieval scenarios. The superior results in MAP, precision-recall curves and Precision within Hamming radius 2 curves suggest that TAH achieves the state of the art performance for search within Hamming Radius 2 on conventional image retrieval problems where the training set and the database are from the same domain. 

\textbf{VisDA2017:} The MAP results of all methods are compared in Table~\ref{table:MAP}. We can observe that for novel transfer retrieval tasks between two domains of VisDA2017, TAH outperforms the comparison methods on the two transfer tasks by very large margins. In particular, compared to the best deep hashing method HashNet, TAH achieves absolute increases of \textbf{0.073} and \textbf{0.090} on the transfer retrieval tasks $synthetic \rightarrow real$ and $real \rightarrow synthetic$ respectively, validating the importance of mitigating domain gap in the transfer setting. Futhermore, compared to state of the art cross-domain deep hashing method THN, we achieve absolute increases of \textbf{0.140} and \textbf{0.096} in average MAP on the transfer retrieval tasks $synthetic \rightarrow real$ and $real \rightarrow synthetic$ respectively. This indicates that the our adversarial learning module is superior to MMD used in THN in aligning distributions. Similarly, the precision-recall curves within Hamming Radius 2 based on 64-bits hash codes for the two transfer retrieval tasks in Figure~\ref{fig:pr_train}-\ref{fig:pr_validation} show that TAH achieves the highest precision at all recall levels. From the Precision within Hamming radius 2 curves shown in Figure~\ref{fig:ham_train}-\ref{fig:ham_validation}, we can observe that TAH outperforms other methods at different bits and has only a moderate decrease of precision when increasing the code length.

In particular, between two transfer retrieval tasks, TAH outperforms other methods with larger margin on $synthetic \rightarrow real$ task. Because the synthetic images contain less information and noise such as background and color than real images. Thus, directly applying the model trained on synthetic images to the real image task suffers from large domain gap or even fail. Transferring knowledge is very important in this task, which explains the large improvement from single domain methods to TAH. TAH also outperforms THN, indicating that adversarial network can match the distribution of two domains better than MMD, and the proposed similarity function based on $t$-distribution can better concentrate data points to be within Hamming radius $2$.

An counter-intuitive result is that the precision keeps unchanged while the recall increases, as shown in Figure \ref{fig:pr}. One plausible reason is that, we present a $t$-distribution motivated hashing loss to enable Hamming space retrieval. Our new loss can concentrate as many data points as possible to be within Hamming ball with radius 2. This concentration property naturally leads to stable precision at different recall levels, i.e. the precision decreases much more slowly by increasing the recall. 

\begin{figure}[!tbp]
  \centering
  \includegraphics[width=1.0\columnwidth]{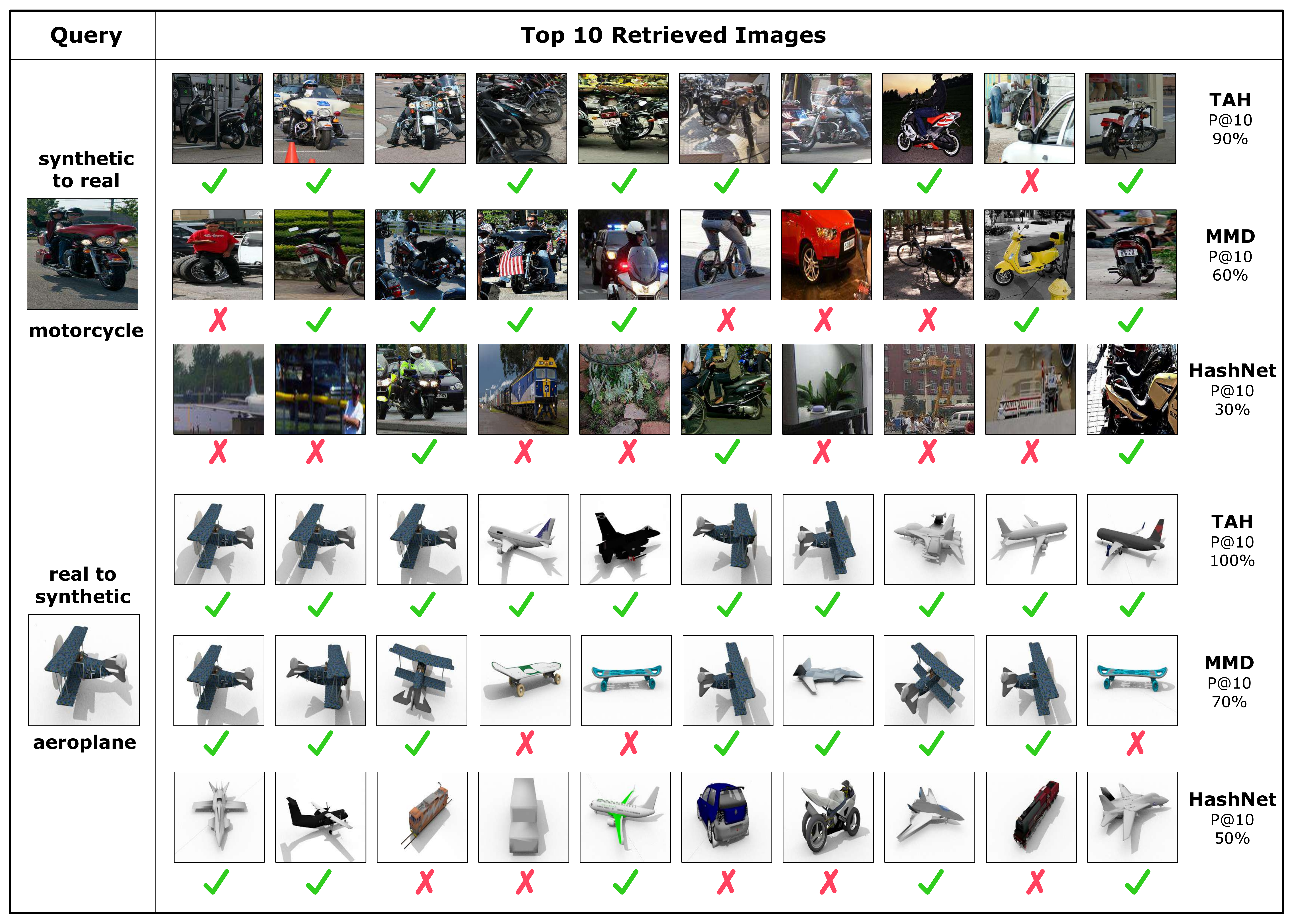}
  \caption{Examples of top 10 retrieved images and P@10.}
   \label{fig:top10}
   \vspace{-10pt}
\end{figure}

Furthermore, as an intuitive illustration, we visualize the top 10 relevant images for a query image for TAH, DHN and HashNet on $synthetic \rightarrow real$ and $real \rightarrow synthetic$ tasks in Figure~\ref{fig:top10}. It shows that TAH can yield much more relevant and user-desired retrieval results.

The superior results of MAP, precision-recall curves and precision within Hamming Radius 2 suggest that TAH is a powerful approach to for learning transferable hash codes for image retrieval. TAH integrates similarity relationship learning and domain adversarial learning into an end-to-end hybrid deep architecture to build the relationship between database points. The results on the NUS-WIDE dataset already show that the similarity relationship learning module is effective to preserve similarity between hash codes and concentrate hash codes of similar points. The experiment on the VisDA2017 dataset further validates that the domain adversarial learning between the source and target domain contributes significantly to the retrieval performance of TAH on transfer retrieval tasks. Since the training and the database sets are collected from different domains and follow different data distributions, there is a substantial domain gap posing a major difficulty to bridge them. The domain adversarial learning module of TAH effectively close the domain gap by matching data distributions with adversarial network. This makes the proposed TAH a good fit for the transfer retrieval.

\vspace{-10pt}
\begin{table}[htp]
%    \addtolength{\tabcolsep}{-2pt} 
    \centering 
    \caption{MAP within Hamming Radius 2 of TAH variants}
    \label{table:EmpirMAP}
	  \resizebox{\columnwidth}{!}{
    \begin{tabular}{c|cccc|cccc}
        \Xhline{1.0pt}
        \multirow{2}{30pt}{\centering Method} & \multicolumn{4}{c|}{\centering $synthetic \rightarrow real$} &\multicolumn{4}{c}{\centering $real \rightarrow synthetic$} \\
        \cline{2-9}
        & 16 bits & 32 bits  & 48 bits  & 64 bits & 16 bits & 32 bits  & 48 bits  & 64 bits\\
        \hline
         TAH-t & \underline{0.443}  & \underline{0.405} & \underline{0.390} & 0.\underline{364} & \underline{0.660} & 0.671 & 0.717 & 0.624 \\
         TAH-A & 0.305  & 0.395 & 0.382 & 0.331 & 0.605 & \underline{0.683} & 0.\underline{725} & \underline{0.724} \\
         TAH & \textbf{0.465} & \textbf{0.423}  & \textbf{0.433} & \textbf{0.404} &\textbf{0.672}  & \textbf{0.695}& \textbf{0.784} & \textbf{0.761}  \\
        \Xhline{1.0pt}
    \end{tabular}
    }
    \vspace{-10pt}
\end{table}

\subsection{Discussion}
We investigate the variants of TAH on VisDA2017 dataset: (1) \textbf{TAH-t} is the variant which uses the pairwise cross-entropy loss introduced in DHN~\cite{cite:AAAI16DHN} instead of our pairwise $t$-distribution cross-entropy loss; (2) \textbf{TAH-A} is the variant removing adversarial learning module and trained without using the unsupervised training data. We report the MAP within Hamming Radius 2 results of all TAH variants on VisDA2017 in Table~\ref{table:EmpirMAP}, which reveal the following observations. (1) TAH outperforms TAH-t by very large margins of 0.031 / 0.060 in average MAP, which confirms that the pairwise $t$ cross-entropy loss learns codes within Hamming Radius 2 better than pairwise cross-entropy loss. (2) TAH outperforms TAH-A by 0.078 / 0.044 in average MAP for transfer retrieval tasks $synthetic \rightarrow real$ and $real \rightarrow synthetic$. This convinces that TAH can further exploit the unsupervised train data of target domain to bridge the Hamming spaces of training dataset (real/synthetic) and database (synthetic/real) and transfer knowledge from training set to database effectively. 

\section{Conclusion}
In this paper, we have formally defined a new transfer hashing problem for image retrieval, and proposed a novel transfer adversarial hashing approach based on a hybrid deep architecture. The key to this transfer retrieval problem is to align different domains in Hamming space and concentrate the hash codes to be within a small Hamming ball, which relies on relationship learning and distribution alignment. Empirical results on public image datasets show the proposed approach yields state of the art image retrieval performance.

\section*{Acknowledgments}
This work was supported by the National Key Research and Development Program of China (2016YFB1000701), National Natural Science Foundation of China (61772299, 61325008, 61502265, 61672313) and TNList Fund.

\begin{small}
  \bibliographystyle{aaai}
  \bibliography{tsne}

\begin{thebibliography}{}

\bibitem[\protect\citeauthoryear{Cao \bgroup et al\mbox.\egroup
  }{2016a}]{cite:KDD16DVSH}
Cao, Y.; Long, M.; Wang, J.; Yang, Q.; and Yu, P.~S.
\newblock 2016a.
\newblock Deep visual-semantic hashing for cross-modal retrieval.
\newblock In {\em KDD}.

\bibitem[\protect\citeauthoryear{Cao \bgroup et al\mbox.\egroup
  }{2016b}]{cite:AAAI16DQN}
Cao, Y.; Long, M.; Wang, J.; Zhu, H.; and Wen, Q.
\newblock 2016b.
\newblock Deep quantization network for efficient image retrieval.
\newblock In {\em AAAI}.
\newblock AAAI.

\bibitem[\protect\citeauthoryear{Cao \bgroup et al\mbox.\egroup
  }{2017a}]{cite:AAAI17THN}
Cao, Z.; Long, M.; Wang, J.; and Yang, Q.
\newblock 2017a.
\newblock Transitive hashing network for heterogeneous multimedia retrieval.
\newblock In {\em AAAI},  81--87.

\bibitem[\protect\citeauthoryear{Cao \bgroup et al\mbox.\egroup
  }{2017b}]{cite:ICCV17HashNet}
Cao, Z.; Long, M.; Wang, J.; and Yu, P.~S.
\newblock 2017b.
\newblock Hashnet: Deep learning to hash by continuation.
\newblock In {\em ICCV}.

\bibitem[\protect\citeauthoryear{Donahue \bgroup et al\mbox.\egroup
  }{2014}]{cite:ICML14DeCAF}
Donahue, J.; Jia, Y.; Vinyals, O.; Hoffman, J.; Zhang, N.; Tzeng, E.; and
  Darrell, T.
\newblock 2014.
\newblock Decaf: A deep convolutional activation feature for generic visual
  recognition.
\newblock In {\em ICML}.

\bibitem[\protect\citeauthoryear{Erin~Liong \bgroup et al\mbox.\egroup
  }{2015}]{cite:CVPR15DH}
Erin~Liong, V.; Lu, J.; Wang, G.; Moulin, P.; and Zhou, J.
\newblock 2015.
\newblock Deep hashing for compact binary codes learning.
\newblock In {\em CVPR},  2475--2483.
\newblock IEEE.

\bibitem[\protect\citeauthoryear{Fleet, Punjani, and
  Norouzi}{2012}]{cite:CVPR12MIH}
Fleet, D.~J.; Punjani, A.; and Norouzi, M.
\newblock 2012.
\newblock Fast search in hamming space with multi-index hashing.
\newblock In {\em CVPR}.
\newblock IEEE.

\bibitem[\protect\citeauthoryear{Ganin and
  Lempitsky}{2015}]{cite:ICML15RevGrad}
Ganin, Y., and Lempitsky, V.
\newblock 2015.
\newblock Unsupervised domain adaptation by backpropagation.
\newblock In {\em ICML}.

\bibitem[\protect\citeauthoryear{Gionis \bgroup et al\mbox.\egroup
  }{1999}]{cite:VLDB99LSH}
Gionis, A.; Indyk, P.; Motwani, R.; et~al.
\newblock 1999.
\newblock Similarity search in high dimensions via hashing.
\newblock In {\em VLDB}, volume~99,  518--529.
\newblock ACM.

\bibitem[\protect\citeauthoryear{Gong and Lazebnik}{2011}]{cite:CVPR11ITQ}
Gong, Y., and Lazebnik, S.
\newblock 2011.
\newblock Iterative quantization: A procrustean approach to learning binary
  codes.
\newblock In {\em CVPR},  817--824.

\bibitem[\protect\citeauthoryear{Gong \bgroup et al\mbox.\egroup
  }{2013}]{cite:CVPR13BP}
Gong, Y.; Kumar, S.; Rowley, H.; Lazebnik, S.; et~al.
\newblock 2013.
\newblock Learning binary codes for high-dimensional data using bilinear
  projections.
\newblock In {\em CVPR},  484--491.
\newblock IEEE.

\bibitem[\protect\citeauthoryear{He \bgroup et al\mbox.\egroup
  }{2016}]{cite:CVPR16DRL}
He, K.; Zhang, X.; Ren, S.; and Sun, J.
\newblock 2016.
\newblock Deep residual learning for image recognition.
\newblock {\em CVPR}.

\bibitem[\protect\citeauthoryear{Indyk and Motwani}{1998}]{cite:STOC98ANN}
Indyk, P., and Motwani, R.
\newblock 1998.
\newblock Approximate nearest neighbors: Towards removing the curse of
  dimensionality.
\newblock In {\em STOC},  604--613.
\newblock New York, NY, USA: ACM.

\bibitem[\protect\citeauthoryear{Jegou, Douze, and
  Schmid}{2011}]{cite:TPAMI11PQ}
Jegou, H.; Douze, M.; and Schmid, C.
\newblock 2011.
\newblock Product quantization for nearest neighbor search.
\newblock {\em TPAMI} 33(1):117--128.

\bibitem[\protect\citeauthoryear{Jia \bgroup et al\mbox.\egroup
  }{2014}]{cite:MM14Caffe}
Jia, Y.; Shelhamer, E.; Donahue, J.; Karayev, S.; Long, J.; Girshick, R.;
  Guadarrama, S.; and Darrell, T.
\newblock 2014.
\newblock Caffe: Convolutional architecture for fast feature embedding.
\newblock In {\em ACM MM}.
\newblock ACM.

\bibitem[\protect\citeauthoryear{Krizhevsky, Sutskever, and
  Hinton}{2012}]{cite:NIPS12CNN}
Krizhevsky, A.; Sutskever, I.; and Hinton, G.~E.
\newblock 2012.
\newblock Imagenet classification with deep convolutional neural networks.
\newblock In {\em NIPS}.

\bibitem[\protect\citeauthoryear{Kulis and Darrell}{2009}]{cite:NIPS09BRE}
Kulis, B., and Darrell, T.
\newblock 2009.
\newblock Learning to hash with binary reconstructive embeddings.
\newblock In {\em NIPS},  1042--1050.

\bibitem[\protect\citeauthoryear{Lai \bgroup et al\mbox.\egroup
  }{2015}]{cite:CVPR15DNNH}
Lai, H.; Pan, Y.; Liu, Y.; and Yan, S.
\newblock 2015.
\newblock Simultaneous feature learning and hash coding with deep neural
  networks.
\newblock In {\em CVPR}.

\bibitem[\protect\citeauthoryear{Lew \bgroup et al\mbox.\egroup
  }{2006}]{cite:TOMM06CBIR}
Lew, M.~S.; Sebe, N.; Djeraba, C.; and Jain, R.
\newblock 2006.
\newblock Content-based multimedia information retrieval: State of the art and
  challenges.
\newblock {\em TOMM} 2(1):1--19.

\bibitem[\protect\citeauthoryear{Li, Wang, and Kang}{2016}]{cite:IJCAI16DPSH}
Li, W.-J.; Wang, S.; and Kang, W.-C.
\newblock 2016.
\newblock Feature learning based deep supervised hashing with pairwise labels.
\newblock In {\em IJCAI}.

\bibitem[\protect\citeauthoryear{Liu \bgroup et al\mbox.\egroup
  }{2011}]{cite:ICML11AGH}
Liu, W.; Wang, J.; Kumar, S.; and Chang, S.-F.
\newblock 2011.
\newblock Hashing with graphs.
\newblock In {\em ICML}.
\newblock ACM.

\bibitem[\protect\citeauthoryear{Liu \bgroup et al\mbox.\egroup
  }{2012}]{cite:CVPR12KSH}
Liu, W.; Wang, J.; Ji, R.; Jiang, Y.-G.; and Chang, S.-F.
\newblock 2012.
\newblock Supervised hashing with kernels.
\newblock In {\em CVPR}.
\newblock IEEE.

\bibitem[\protect\citeauthoryear{Liu \bgroup et al\mbox.\egroup
  }{2013}]{cite:CVPR13HBS}
Liu, X.; He, J.; Lang, B.; and Chang, S.-F.
\newblock 2013.
\newblock Hash bit selection: a unified solution for selection problems in
  hashing.
\newblock In {\em CVPR},  1570--1577.
\newblock IEEE.

\bibitem[\protect\citeauthoryear{Liu \bgroup et al\mbox.\egroup
  }{2014}]{cite:CVPR14CH}
Liu, X.; He, J.; Deng, C.; and Lang, B.
\newblock 2014.
\newblock Collaborative hashing.
\newblock In {\em CVPR},  2139--2146.

\bibitem[\protect\citeauthoryear{Liu \bgroup et al\mbox.\egroup
  }{2016}]{cite:CVPR2016DSH}
Liu, H.; Wang, R.; Shan, S.; and Chen, X.
\newblock 2016.
\newblock Deep supervised hashing for fast image retrieval.
\newblock In {\em CVPR},  2064--2072.

\bibitem[\protect\citeauthoryear{Norouzi and Blei}{2011}]{cite:ICML11MLH}
Norouzi, M., and Blei, D.~M.
\newblock 2011.
\newblock Minimal loss hashing for compact binary codes.
\newblock In {\em ICML},  353--360.
\newblock ACM.

\bibitem[\protect\citeauthoryear{Norouzi, Blei, and
  Salakhutdinov}{2012}]{cite:NIPS12HDML}
Norouzi, M.; Blei, D.~M.; and Salakhutdinov, R.~R.
\newblock 2012.
\newblock Hamming distance metric learning.
\newblock In {\em NIPS},  1061--1069.

\bibitem[\protect\citeauthoryear{Norouzi, Punjani, and
  Fleet}{2014}]{cite:TPAMI14FES}
Norouzi, M.; Punjani, A.; and Fleet, D.~J.
\newblock 2014.
\newblock Fast exact search in hamming space with multi-index hashing.
\newblock {\em TPAMI} 36(6):1107--1119.

\bibitem[\protect\citeauthoryear{Salakhutdinov and
  Hinton}{2007}]{cite:AI07SemanticHashing}
Salakhutdinov, R., and Hinton, G.~E.
\newblock 2007.
\newblock Learning a nonlinear embedding by preserving class neighbourhood
  structure.
\newblock In {\em AISTATS},  412--419.

\bibitem[\protect\citeauthoryear{Shen \bgroup et al\mbox.\egroup
  }{2015}]{cite:CVPR15SDH}
Shen, F.; Shen, C.; Liu, W.; and Tao~Shen, H.
\newblock 2015.
\newblock Supervised discrete hashing.
\newblock In {\em CVPR}.
\newblock IEEE.

\bibitem[\protect\citeauthoryear{Smeulders \bgroup et al\mbox.\egroup
  }{2000}]{cite:TPAMI00SemanticGap}
Smeulders, A.~W.; Worring, M.; Santini, S.; Gupta, A.; and Jain, R.
\newblock 2000.
\newblock Content-based image retrieval at the end of the early years.
\newblock {\em TPAMI} 22(12):1349--1380.

\bibitem[\protect\citeauthoryear{Wang \bgroup et al\mbox.\egroup
  }{2014}]{cite:Arxiv14HashSurvey}
Wang, J.; Shen, H.~T.; Song, J.; and Ji, J.
\newblock 2014.
\newblock Hashing for similarity search: A survey.
\newblock Arxiv.

\bibitem[\protect\citeauthoryear{Wang, Kumar, and
  Chang}{2012}]{cite:TPAMI12SSH}
Wang, J.; Kumar, S.; and Chang, S.-F.
\newblock 2012.
\newblock Semi-supervised hashing for large-scale search.
\newblock {\em TPAMI} 34(12):2393--2406.

\bibitem[\protect\citeauthoryear{Weiss, Torralba, and
  Fergus}{2009}]{cite:NIPS09SH}
Weiss, Y.; Torralba, A.; and Fergus, R.
\newblock 2009.
\newblock Spectral hashing.
\newblock In {\em NIPS}.

\bibitem[\protect\citeauthoryear{Xia \bgroup et al\mbox.\egroup
  }{2014}]{cite:AAAI14CNNH}
Xia, R.; Pan, Y.; Lai, H.; Liu, C.; and Yan, S.
\newblock 2014.
\newblock Supervised hashing for image retrieval via image representation
  learning.
\newblock In {\em AAAI},  2156--2162.
\newblock AAAI.

\bibitem[\protect\citeauthoryear{Yang \bgroup et al\mbox.\egroup
  }{2017}]{cite:AAAI17PRDH}
Yang, E.; Deng, C.; Liu, W.; Liu, X.; Tao, D.; and Gao, X.
\newblock 2017.
\newblock Pairwise relationship guided deep hashing for cross-modal retrieval.

\bibitem[\protect\citeauthoryear{Yu \bgroup et al\mbox.\egroup
  }{2014}]{cite:ICML14CBE}
Yu, F.~X.; Kumar, S.; Gong, Y.; and Chang, S.-F.
\newblock 2014.
\newblock Circulant binary embedding.
\newblock In {\em ICML},  353--360.
\newblock ACM.

\bibitem[\protect\citeauthoryear{Zhang \bgroup et al\mbox.\egroup
  }{2014}]{cite:SIGIR14LFH}
Zhang, P.; Zhang, W.; Li, W.-J.; and Guo, M.
\newblock 2014.
\newblock Supervised hashing with latent factor models.
\newblock In {\em SIGIR},  173--182.
\newblock ACM.

\bibitem[\protect\citeauthoryear{Zhu \bgroup et al\mbox.\egroup
  }{2016}]{cite:AAAI16DHN}
Zhu, H.; Long, M.; Wang, J.; and Cao, Y.
\newblock 2016.
\newblock Deep hashing network for efficient similarity retrieval.
\newblock In {\em AAAI}.
\newblock AAAI.

\end{thebibliography}
\end{small}

\end{document}